\documentclass[letterpaper, 10 pt, conference]{ieeeconf}  

\IEEEoverridecommandlockouts                              

\title{\LARGE \bf
VP-STO: Via-point-based Stochastic Trajectory Optimization for Reactive Robot Behavior
}

\author{Julius Jankowski$^{* \; 1,2}$, Lara Bruderm\"uller$^{* \; 3}$, Nick Hawes$^{3}$ and Sylvain Calinon$^{1,2}$
\thanks{*Authors contributed equally.}
\thanks{JJ and SC were supported by the Swiss National Science Foundation (SNSF) through the CODIMAN project. LB was supported by an Amazon Web Services Lighthouse scholarship. NH received EPSRC funding via the ``From Sensing to Collaboration'' programme grant [EP/V000748/1].
}%
\thanks{$^{1}$Idiap Research Institute, Martigny,
CH; {\tt\footnotesize name.surname@idiap.ch}}%
\thanks{$^{2}$Ecole Polytechnique Fédérale de Lausanne (EPFL), CH}%
\thanks{$^{3}$Oxford Robotics Institute, University of Oxford, UK; {\tt\footnotesize \{larab, nickh\}@robots.ox.ac.uk.}}%
}

\usepackage{amsmath, amssymb, bm, algorithmic, algorithm2e, xcolor, xargs}
\usepackage{todonotes}
\usepackage{hyperref}
\hypersetup{
    colorlinks=true,
    linkcolor=black,
    filecolor=magenta,   
    urlcolor=blue,
    }

\SetCommentSty{mycommfont}

\newcommand{\trsp}{{\scriptscriptstyle\top}}

\newcommandx{\julius}[2][1=]{\todo[backgroundcolor=blue!25,inline,#1]{JJ: #2}\noindent}
\newcommandx{\lara}[2][1=]{\todo[backgroundcolor=green!25,inline,#1]{LB: #2}\noindent}
\newcommandx{\sylvain}[2][1=]{\todo[backgroundcolor=red!25,inline,#1]{SC: #2}\noindent}

\newcommand{\eg}[0]{\textit{e.g.}, }
\newcommand{\ie}[0]{\textit{i.e.}, }

\begin{document}
\maketitle
\thispagestyle{empty}
\pagestyle{empty}


\begin{abstract}
Achieving reactive robot behavior in complex dynamic environments is still challenging as it relies on being able to solve trajectory optimization problems quickly enough, such that we can replan the future motion at frequencies which are sufficiently high for the task at hand. We argue that current limitations in Model Predictive Control (MPC) for robot manipulators arise from inefficient, high-dimensional trajectory representations and the negligence of time-optimality in the trajectory optimization process. Therefore, we propose a motion optimization framework that optimizes \emph{jointly} over space and time, generating smooth and timing-optimal robot trajectories in joint-space. While being task-agnostic, our formulation can incorporate additional task-specific requirements, such as collision avoidance, and yet maintain real-time control rates, demonstrated in simulation and real-world robot experiments on closed-loop manipulation. For additional material, please visit \url{https://sites.google.com/oxfordrobotics.institute/vp-sto}.
\end{abstract}


\section{Introduction}

In this paper we consider the problem of generating continuous, \emph{timing-optimal} and smooth trajectories for robots operating in dynamic environments. Such task settings require the robot to be \emph{reactive} to unforeseen changes in the environment, \eg due to dynamic obstacles, as well as to be \emph{robust} and \emph{compliant} when operating alongside or together with humans. 
However, generating this kind of reactive and yet efficient robot behavior within a high-dimensional configuration space is significantly challenging. This is especially the case in robot manipulation scenarios with many degrees of freedom (DoFs) as the resulting high-dimensional and multi-objective optimization problems are difficult to solve on-the-fly. 
%
A widespread approach in robotics is to formulate the task of motion generation as an optimization problem. Such \emph{trajectory-optimization} based methods aim at finding a trajectory that minimizes a cost function, \eg motion smoothness, subject to constraints, \eg collision avoidance. 
%
Solution strategies can either be \emph{gradient-based} or \emph{sampling-based}. Approaches falling in the former category, \eg CHOMP \cite{Zucker2013} and \mbox{TrajOpt} \cite{Schulman2014}, typically employ second-order iterative methods to find locally optimal solutions. However, they require the cost function to be once or even twice-differentiable, which constitutes a major limitation for manipulation tasks as they usually involve many complex, discontinuous cost terms and constraints. In contrast, sampling-based methods \cite{Kalakrishnan2011, Rubinstein1992} can operate on discontinuous costs by sampling candidate trajectories from a proposal distribution, evaluating them on the objective, and updating the proposal distribution according to their relative performance. Compared to gradient-based optimization, stochastic approaches typically also achieve higher robustness to difficult reward landscapes due to their exploratory properties \cite{Hansen2016}. 
%
\begin{figure}[t]
    \centering
    \includegraphics[width=\linewidth]{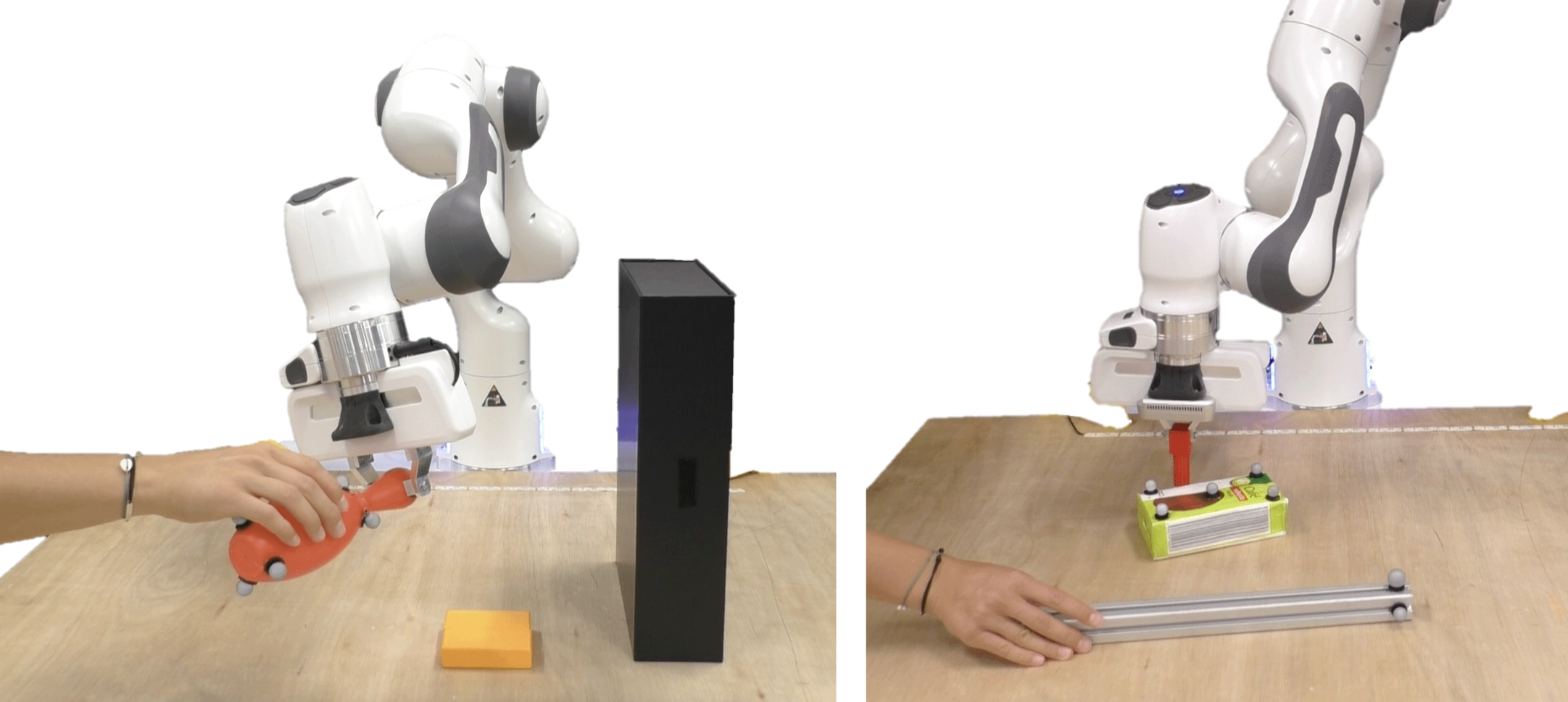}
    \caption{\emph{\textbf{Experiment settings.}} \emph{Left:} Pick-and-place scenario, where the task is to grasp a bowling pin that is arbitrarily handed over to the robot and to place it upright in the middle of the table. \emph{Right:} Pushing scenario, where the robot has to push the center of the green coffee packet to a moving target location indicated by the tip of the metal stick.}
    \label{fig:panda_exp}
    \vskip -19pt
\end{figure}
Yet, achieving reactive robot behavior is challenging as it requires solving trajectory optimization problems at frequencies which are sufficiently high for the task at hand. 
This issue can be alleviated in \emph{Model Predictive Control (MPC)} settings by optimizing over a shorter receding time-horizon. Stochastic, gradient-free trajectory optimization, such as Model-Predictive Path Integral (MPPI) control \cite{Williams2017} and the Cross-Entropy-Method (CEM) \cite{Rubinstein1992}, combined with MPC, also known as \emph{sampling-based MPC}, has proven state-of-the-art real-time performance on real robotic systems in challenging and dynamic environments \cite{Williams2015, bangura_real-time_2014, Bhardwaj2021}. 
However, these works still suffer from limited long-term anticipation, \eg getting stuck in front of obstacles, due to the optimization over a short receding horizon.

Motivated by the above, we propose \emph{Via-Point-based Stochastic Trajectory Optimization (VP-STO)}, a framework that introduces the following contributions 
\begin{enumerate}
    \item A low-dimensional, time-continuous representation of trajectories in joint-space based on via-points that by-design respect kinodynamic constraints of the robot.
    \item Stochastic via-point optimization, based on an evolutionary strategy, aiming at minimizing movement duration and task-related cost terms.
    \item An MPC algorithm optimizing over the full horizon for real-time application in complex high-dimensional task settings, such as closed-loop object manipulation.
\end{enumerate}

\section{Related Work}
In the context of closed-loop object manipulation with MPC, 
successful approaches to producing reactive robot behavior typically optimize in joint-space subject to kinodynamic constraints. 
While Fishman et al. use gradient-based MPC in order to find trajectories for human-robot handovers \cite{Fishman2020}, a very recent approach named STORM \cite{Bhardwaj2021} employed sampling-based MPC on robotic manipulation tasks. It is able to generate particularly smooth trajectories via low discrepancy action sampling, smooth interpolation and careful cost function design. Moreover, the parallelizability of sampling-based MPC is exploited by deploying the stochastic tensor optimization framework on a GPU. 
However, in contrast to our work, the approach relies on optimizing over a short receding horizon.


In the realm of time-parametrization of trajectories, most existing approaches fix the overall motion duration or do not specify it at all. For instance, the majority of MPC-based approaches only handle time implicitly via kinodynamic constraints. 
While the works of \cite{van_den_broeck_model_2011, rosmann_time-optimal_2017} progress the state of the art in time-optimal MPC, their applicability to high-dimensional robotic systems yet is limited.
In the context of motion planning, T-CHOMP \cite{byravan_space-time_2014} jointly optimizes a trajectory and the corresponding via-point timings. Yet, the total execution time is still fixed in advance. 
The way we approach the minimization of the movement duration is most similar to the work of \cite{Toussaint2022}. However, in contrast to our work, their approach optimizes via-points and their timing separately.   

\section{Preliminaries: Trajectory Representation}
\label{sec:obf}
The way we represent trajectories is based on previous work showing that the closed-form solution to the following optimization problem
%
\begin{align}
\begin{split}
\label{eq:obf}
  &\textrm{min} \quad \int_0^{1} \bm{q}''(s)^\trsp \bm{q}''(s) ds \\
  \mathrm{s.t.} \quad &\bm{q}(s_n) = \bm{q}_n, \quad n = 1, ...,N\\
  &\bm{q}(0) = \bm{q}_0, \bm{q}'(0) = \bm{q}'_0, \; \bm{q}(1) = \bm{q}_T, \bm{q}'(1) = \bm{q}'_T
\end{split}
\end{align}
is given by cubic splines \cite{Zhang97} and that it can be formulated as a weighted superposition of basis functions \cite{Jankowski2022}. 
Hence, the robot's configuration is defined as $\bm{q}(s) = \bm{\Phi}(s) \bm{w} \in \mathbb{R}^D$, with $D$ being the number of degrees of freedom.
The matrix $\bm{\Phi}(s)$ contains the basis functions which are weighted by the vector $\bm{w}$\footnote{A more detailed explanation of the basis functions and their derivation can be found in the appendix of \cite{Jankowski2022}.}. 
The trajectory is defined on the interval $\mathcal{S} =  [0,1]$, while the time $t$ maps to the phase variable $s = \frac{t}{T} \in \mathcal{S}$ with $T$ being the total duration of the trajectory.
Consequently, joint velocities and accelerations along the trajectory are given by $\dot{\bm{q}}(s) = \frac{1}{T} \bm{\Phi}'(s) \bm{w}$ and $\ddot{\bm{q}}(s) = \frac{1}{T^2} \bm{\Phi}''(s) \bm{w}$, respectively\footnote{We use the notation $f'(s)$ for derivatives w.r.t. $s$ and the notation $\dot{f}(s)$ for derivatives w.r.t. $t$.}.
%
The basis function weights $\bm{w}$ include the trajectory constraints consisting of the boundary condition parameters $\bm{w}_{\textrm{bc}} = \left[ \bm{q}_0^\trsp, \bm{q}'^\trsp_0, \bm{q}_T^\trsp, \bm{q}'^\trsp_T \right]^\trsp$ and $N$ via-points the trajectory has to pass through $\bm{q}_{\textrm{via}} = \left[ \bm{q}_1^\trsp, \hdots, \bm{q}_N^\trsp \right]^\trsp \in \mathbb{R}^{D N}$, such that $\bm{w} = \left[ \bm{q}_{\textrm{via}}^\trsp, \bm{w}_{\textrm{bc}}^\trsp \right]^\trsp$. 
Throughout this paper, the via-point timings $s_n$ are assumed to be uniformly distributed in $\mathcal{S}$.
Note that boundary velocities map to boundary derivatives w.r.t. $s$ by multiplying them with the total duration $T$, \ie $\bm{q}'_0 = T \dot{\bm{q}}_0$ and $\bm{q}'_T = T \dot{\bm{q}}_T$.
Furthermore, the optimization problem in Eq.~\eqref{eq:obf} minimizes not only the objective $\bm{q}''(s)$, but also the integral over accelerations, since $\bm{q}''(s) = T^2 \ddot{\bm{q}}(s)$ and thus the objective $\int_0^{1} \ddot{\bm{q}}(s)^\trsp \ddot{\bm{q}}(s) ds$ directly maps to $\frac{1}{T^4} \int_0^{1} \bm{q}''(s)^\trsp \bm{q}''(s) ds$, corresponding to the control effort. It is minimal \textit{iff} the objective in Eq.~\eqref{eq:obf} is minimal. 
As a result, this trajectory representation provides a linear mapping from via points, boundary conditions and the movement duration to a time-continuous and smooth trajectory.

\begin{figure*}[ht!]
    \centering
    \includegraphics[width=\linewidth]{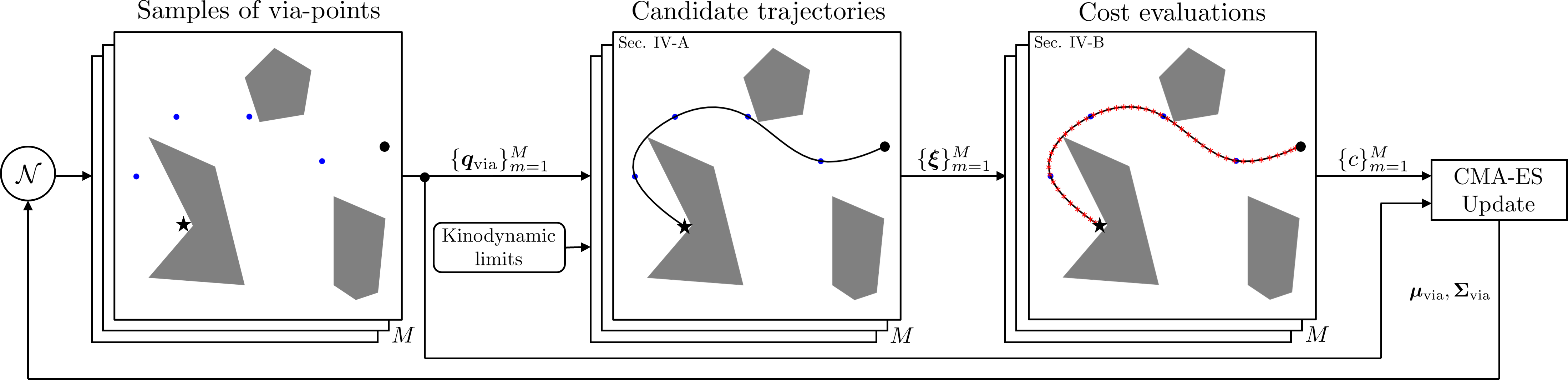}
    \caption{An illustration of the via-point-based stochastic trajectory optimization loop. First, a new population of $M$ via-points $\bm{q}_{\textrm{via}}$ is sampled from a Gaussian distribution $\mathcal{N}(\bm{\mu}_{\textrm{via}},\bm{\Sigma}_{\textrm{via}})$. Then, the sampled via-points are transformed into a population of candidate trajectories subject to kinodynamic limits. Next, the resulting trajectories are ranked according to their cost evaluations. Last, the parameters of the Gaussian sampling distribution are updated via CMA-ES using the cost rankings and the via-point sets themselves.}
    \label{fig:vpsto_opt}
    \vskip -16pt
\end{figure*}

In the remainder of the paper, we exploit this explicit parameterization with via-points and boundary conditions by optimizing only the via-points while keeping the predefined boundary condition parameters fixed.
Thus, we write the computation of the trajectory as a superposition of a via-point term and a boundary constraints term, \ie
%
    $\bm{q}(s) = \bm{\Phi}_{\textrm{via}}(s) \bm{q}_{\textrm{via}} + \bm{\Phi}_{\textrm{bc}}(s) \bm{w}_{\textrm{bc}}$.
%
The matrices $\bm{\Phi}_{\textrm{via}}(s)$ and $\bm{\Phi}_{\textrm{bc}}(s)$ are extracted from the basis function matrix $\bm{\Phi}(s)$.

\section{VP-Sto: Via-Point-based Stochastic Trajectory Optimization}
\label{sec:vpsto_offline}

In the following, we introduce our stochastic trajectory optimization framework.
The core idea is to find via-points $\bm{q}_{\textrm{via}}$ such that the synthesized trajectory minimizes a task-related objective, \ie
\begin{equation}
    \min_{\bm{q}_{\textrm{via}}} \quad c \left[ \, \bm{q}(s), \dot{\bm{q}}(s), \ddot{\bm{q}}(s), T \, \right].
    \label{eq:cost_func}
\end{equation}
Based on these via-points, we efficiently synthesize high-quality trajectories, \ie $\bm{q}_{\textrm{via}} \rightarrow \bm{\xi}$ with $\bm{\xi} = \left\{\bm{q}(s), \dot{\bm{q}}(s), \ddot{\bm{q}}(s), T\right\}$.
We aim at synthesizing trajectories that \textit{by-design} minimize task-agnostic objectives, \ie \textit{minimum time} and \textit{smoothness}, and satisfy task-agnostic constraints, \ie equality constraints on the \textit{initial} and \textit{final state} and inequality constraints on \textit{joint-space velocities} and \textit{accelerations}.
We employ stochastic black-box optimization, namely \emph{Covariance Matrix Adaptation (CMA-ES)} \cite{Hansen2016} to optimize for the via-points. As each trajectory constructed from the sampled via-points already provides the optimal solution to the optimization problem given in Eq.~\ref{eq:obf}, the CMA-ES optimization in the low-dimensional via-point space is particularly fast, evaluating only high-quality trajectories. 
Moreover, with CMA-ES we are not only able to quickly converge to a local minimum, but to also leverage the exploration aspect of the evolutionary strategy (ES). In more detail, this nested optimization process, which is also illustrated in Fig.~\ref{fig:vpsto_opt}, comprises the following steps. First, a new population of $M$ via-points $\bm{q}_{\textrm{via}}$ is sampled from a Gaussian distribution $\mathcal{N}(\bm{\mu}_{\textrm{via}},\bm{\Sigma}_{\textrm{via}})$. As $\bm{q}_{\textrm{via}}$ is a vector of the stacked via-points, note that $\bm{\mu}_{\textrm{via}} \in \mathbb{R}^{D N}$ and $\bm{\Sigma}_{\textrm{via}} \in \mathbb{R}^{D N \times D N}$. By taking $M$ samples in this higher-dimensional space, instead of $M\cdot N$ samples for all via points separately in the configuration space, we are able to sample $M$ sets of correlated via-points. 
Then, as described in detail in Sec.~\ref{sec:synthesis}, the sampled via-points are transformed into a population of candidate trajectories that are evaluated according to cost terms as outlined in Sec.~\ref{sec:eval}.
Finally, we use CMA-ES in order to update the parameters $\bm{\mu}_{\textrm{via}}, \bm{\Sigma}_{\textrm{via}}$ of the Gaussian distribution of via-points. This optimization setup enables us to find a valid local minimum or even the global minimum at rates sufficient for reactive robot behavior in closed-loop manipulation tasks, as we demonstrate in our experiments outlined in Section~\ref{sec:exp}.

\subsection{Synthesis of Kinodynamically Admissible Trajectories}
\label{sec:synthesis}

In this section, we show how we translate the sampled via-points $\bm{q}_{\textrm{via}}$ into kinodynamically admissible trajectories. So far, the trajectory is implicitly given in phase space as described in Sec.~\ref{sec:obf}. 
Given the via-points and the boundary conditions $\bm{q}_0, \dot{\bm{q}}_0, \bm{q}_T, \dot{\bm{q}}_T$, the computation of the explicit continuous trajectory only depends on the total movement duration $T$. 
It is determined by the dynamic limits on velocity $\dot{\bm{q}}_{\textrm{min}}, \dot{\bm{q}}_{\textrm{max}}$ and acceleration $\ddot{\bm{q}}_{\textrm{min}}, \ddot{\bm{q}}_{\textrm{max}}$ and is thus given as the minimum positive duration such that the resulting velocity and acceleration profiles satisfy the limits.
%
%
We find a sufficient approximation for $T$ by solving the above problem over a discrete set of $K$ evaluation points $\{s_k\}_{k=0}^K$, uniformly distributed in the continuous phase space $\mathcal{S}$. For each evaluation point there exists a closed-form solution $T_k(\bm{q}_{\textrm{via}})$ that is computationally cheap to evaluate.
We then pick the most conservative duration among the $K$ solutions for $T$, \ie $T(\bm{q}_{\textrm{via}}) = \max_k T_k(\bm{q}_{\textrm{via}})$, in order to make sure that the velocity and acceleration constraints are satisfied at all evaluation points.
This procedure will result in trajectories where either the velocity or the acceleration profile reaches the limit in at least one evaluation point.
Finally, having determined $T$, we are able to explicitly compute the kinodynamically admissible trajectory $\bm{\xi}$.


\subsection{Cost Evaluation}
\label{sec:eval}

Given the sampled and synthesized population of trajectories, we evaluate the performance, \ie the cost $c$ of each trajectory, independently.
The gradient-free optimizer allows for sharp cost function profiles, \eg trajectory constraints expressed through discontinuous barrier functions (cf. Sec.~\ref{sec:exp} for examples).
We approximate $c(\bm{\xi})$ by sampling the given trajectory with a predefined resolution $\Delta s$ in the phase space $\mathcal{S}$ and accumulating the costs at these $K$ evaluation points.
In the time domain this can still map to varying resolutions of individual trajectories, as $\Delta t = T \Delta s$. 
Note that the evaluation points at $s_k$ are not equivalent to the via-points at $s_n$, as depicted in Fig.~\ref{fig:vpsto_opt}.
The resolution of $s_k$ can be higher than that of $s_n$ in order to have a better approximation of the trajectory cost while keeping the actual optimization variable $\bm{q}_{\textrm{via}}$ low-dimensional.

\section{Online VP-STO (MPC)}\label{sec:vpsto_online}

In order to perform closed-loop control via continuous online re-optimization, we embed the \emph{VP-STO} framework into an MPC algorithm.
In this online setting, the main focus lies on rapidly finding valid movements connecting the current robot state $\bm{q}, \dot{\bm{q}}$ with a goal state $\bm{q}_T, \dot{\bm{q}}_T$ and re-optimizing them at a sufficient rate $f_{\textrm{mpc}} = \frac{1}{\Delta t_{\textrm{mpc}}}$. Algorithm~\ref{alg:mpc} outlines a single MPC step that, given the current robot state, attempts to find an optimal \emph{full-horizon} trajectory and to extract a short-horizon reference to be tracked by a lower-level impedance controller. The details of the algorithm will be outlined in the remainder of this section.

\RestyleAlgo{ruled}
\SetKwComment{Comment}{/* }{ */}
\SetKw{AND}{and}
\SetKw{IN}{Input:}

\begin{algorithm}[hbt!]
\DontPrintSemicolon
\caption{Online VP-STO: $i$-th MPC Step}\label{alg:mpc}
\KwIn{$\bm{q}, \dot{\bm{q}}, \bm{q}_T, \dot{\bm{q}}_T, \dot{\bm{q}}_{\textrm{min}}, \dot{\bm{q}}_{\textrm{max}}, \ddot{\bm{q}}_{\textrm{min}}, \ddot{\bm{q}}_{\textrm{max}}$, $\Delta t_{\textrm{mpc}}$, $T_{\textrm{stop}}$}
\KwOut{Short-horizon reference $\bm{q}_{\textrm{d}}(t), \dot{\bm{q}}_{\textrm{d}}(t), \ddot{\bm{q}}_{\textrm{d}}(t)$}
$t_{\textrm{optimize}} \gets 0$\;
$\bm{q}_0, \dot{\bm{q}}_0 \gets \bm{q}, \dot{\bm{q}}$\;
$\bm{\xi}_{\textrm{direct}} \gets$ synthesize() \tcp*{\ref{sec:direct}}
\eIf{$\bm{\xi}_{\textrm{direct}}$ is valid \AND $\bm{\xi}_{\textrm{direct}}$ is shorter than $T_{\textrm{stop}}$}{
    $\bm{\xi}^*_{i} \gets \bm{\xi}_{\textrm{direct}}$\;
}{
    \eIf{$\bm{\xi}^*_{i-1}$ is valid}{
        ${}^0\bm{\mu}_{\textrm{via}}, {}^0\bm{\Sigma}_{\textrm{via}}, N \gets$ warmStart$\left(\bm{\xi}^*_{i-1}\right)$ \tcp*{\ref{sec:init}}
    }{
        ${}^0\bm{\mu}_{\textrm{via}}, {}^0\bm{\Sigma}_{\textrm{via}}, N \gets$ exploreInit() \tcp*{\ref{sec:init}}
    }
    $j \gets 0$\;
    \While{$t_{\textrm{optimize}} < \Delta t_{\textrm{mpc}}$}{
        $\{\bm{q}_{\textrm{via}}\}_{m=1}^M \gets$ sample$\left({}^j\bm{\mu}_{\textrm{via}}, {}^j\bm{\Sigma}_{\textrm{via}}\right)$ \tcp*{\ref{sec:sampling}}
        $\{\bm{\xi}\}_{m=1}^M \gets$ synthesize$\left(\{\bm{q}_{\textrm{via}}\}_{m=1}^M\right)$ \tcp*{\ref{sec:synthesis}}
        $\{c\}_{m=1}^M \gets$ evaluate$\left(\{\bm{\xi}\}_{m=1}^M\right)$ \tcp*{\ref{sec:eval}}
        $\bm{\mu}^{j+1}_{\textrm{via}}, \bm{\Sigma}^{j+1}_{\textrm{via}} \gets$ sep-CMA-ES$\left(\{\bm{q}_{\textrm{via}}, c\}_{m=1}^M\right)$\;
        $j \gets j+1$\;
    }
    $\bm{\xi}^*_{i} \gets$ synthesize$\left(\bm{\mu}^{j}_{\textrm{via}}\right)$\;
}
$\bm{q}_{\textrm{d}}(t), \dot{\bm{q}}_{\textrm{d}}(t), \ddot{\bm{q}}_{\textrm{d}}(t) \gets$ shortHorizon$\left(\bm{\xi}^*_{i}\right)$ \tcp*{\ref{sec:control}}
\vskip -2pt
\end{algorithm}



In the online setting the number of via-points $N$ used to parameterize the trajectory plays an important role. 
A large number of via-points can capture highly complex movements and may find more optimal solutions.
However, it also implies a higher-dimensional decision space which increases the computational complexity of the optimization loop.
Consequently, a particular focus within the MPC algorithm lies on the selection of $N$.

\subsection{No-Via-Point Trajectory for Stopping Behavior}
\label{sec:direct}

\emph{VP-STO} is based on optimizing the locations of a given number of via-points.
However, the trajectory synthesis, described in Sec.~\ref{sec:synthesis}, also works without any via-points, \ie $N{=}0$.
The resulting trajectory connects the current robot state and the desired state by a third-order polynomial that minimizes the smoothness objective in Eq.~\eqref{eq:obf} and satisfies the kinodynamic limits.
As this no-via-point trajectory is a unique solution, it can not account for any other movement objectives, \eg to avoid collisions. 
Yet, the advantage is a cheap-to-construct trajectory that has no stochasticity, which is useful for driving the robot to the target configuration and stopping with zero velocity.
Therefore, at the beginning of each optimization cycle, we first check if this simple direct trajectory is valid, \eg collision-free, and if the corresponding duration of the movement is below the user-defined threshold $T_{\textrm{stop}}$.
By setting the threshold rather small, we let the mechanism take over towards the final part of the total trajectory to achieve robust stopping behavior for reaching the goal.
If the direct solution is not used, we perform a \textit{VP-STO} optimization cycle.

\subsection{Initialization: Exploration vs. Warm-Starting}
\label{sec:init}

\begin{figure}
\centering
\includegraphics[width=\linewidth]{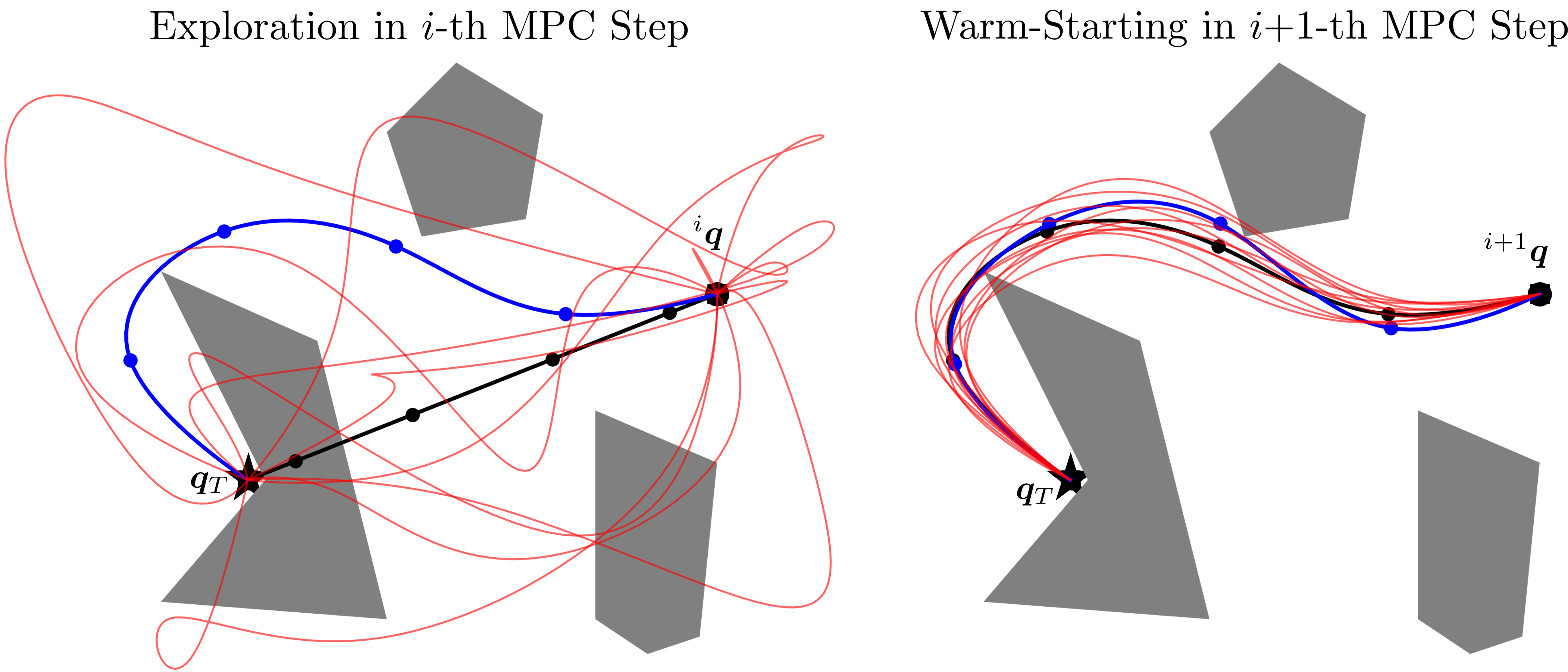}
\caption{An illustration of the stochastic optimization process within the proposed MPC algorithm. \textbf{\emph{Left:}} In the \emph{exploration} mode, trajectories are sampled and synthesized with a large initial variance in order to discover valid solutions. \textbf{\emph{Right:}} If a valid solution is available from the previous MPC step, we \emph{warm-start} the optimization by shifting the solution and sampling from a lower-variance initial distribution. All sampled trajectories are shown in red. The initial guesses ${}^0\bm{\mu}_{\textrm{via}}$ of an MPC step are depicted by the black solid lines, while the blue trajectories illustrate the mean solution ${}^{20}\bm{\mu}_{\textrm{via}}$ after $20$ optimization iterations.}
\label{fig:mpc_init}
\vskip -16pt
\end{figure}

The use of an evolutionary optimization strategy, such as CMA-ES, allows us to initialize the optimization not only with an initial guess of the via-points $\bm{\mu}_{\textrm{via}}$, but also to set the corresponding initial variance $\bm{\Sigma}_{\textrm{via}}$ as an estimate of how certain we are about the initial solution.
The initial variance can thus be interpreted as an exploration parameter influencing how the very first population of candidate trajectories will be sampled. Therefore, in each MPC step we use two possible modes on how to initialize these parameters. The effects of each mode on the resulting candidate trajectories are shown in Fig.~\ref{fig:mpc_init}.

\vspace*{1ex}
\noindent\textbf{Exploration.} If a MPC step was not successful in finding a valid trajectory, the successive MPC step will be used to \textit{explore} a larger area of the trajectory space to ideally discover a valid solution, as can be seen from the sampled trajectories in the left of Fig.~\ref{fig:mpc_init}. 
We initialize the mean solution $\bm{\mu}_{\textrm{via}}$ with a naive straight-line guess with high uncertainty, \ie large diagonal values of $\bm{\Sigma}_{\textrm{via}}$.
The number of via-points used to parameterize the trajectory is set to $N = N_{\max}$. $N_{\max}$ needs to be specified by the user and depends on the complexity of the task, as well as on the available computational resources.

\vspace*{1ex}
\noindent\textbf{Warm-Starting.} If a valid solution was found in a MPC step, we shift the solution forward in time and use it to \textit{warm-start} the mean $\bm{\mu}_{\textrm{via}}$ in the successive MPC step, potentially further improving the current solution.
In this case, we initialize the covariance matrix $\bm{\Sigma}_{\textrm{via}}$ with low values on the diagonal as we are more certain about the proximity of the current solution to a valid local minimum, as can be seen on the right of Fig.~\ref{fig:mpc_init}.
In order to determine the number of via-points $N$ for the successive MPC step, we use the movement duration of the current solution as a proxy for how complex the remainder of the movement will be.
We therefore set $N = \max ( 1, \min (\lceil\alpha T\rceil, N_{\max}))$, where $T$ is the total duration of the current solution and $\alpha$ a user-defined scaling parameter. 

\subsection{Efficient Gaussian Sampling of Smooth Trajectories through Covariance Matrix Decomposition}
\label{sec:sampling}

For the sake of computational efficiency and linear scalability to high-DoF systems in our MPC solver, we use a variant of CMA-ES that iterates on diagonal covariance matrices instead of full covariance matrices, namely \emph{sep-CMA-ES} \cite{rudolph_simple_2008}.
However, a diagonal covariance matrix does not capture the correlations between the sampled via-points that are important for sampling smooth trajectories.
We counteract this disadvantage by using a Cholesky factorization of the covariance matrix, such that $\bm{\Sigma}_{\textrm{via}} = \bm{L} \bm{D} \bm{L}^\trsp$, where the diagonal matrix $\bm{D} = \textrm{diag}(\bm{\sigma}_{\textrm{via}})$ is subject to iterative optimization through sep-CMA-ES. This renders our algorithm to a computational complexity of $\mathcal{O}(ND)$, with $N$ being the number of via-points and $D$ the DoF of the robot, instead of $\mathcal{O}((ND)^2)$ in the case of the full covariance matrix.
The lower triangular matrix $\bm{L}$ is computed offline as the Cholesky decomposition of a constant covariance matrix
\begin{equation}
    \bm{\Sigma}_{\textrm{smooth}} = \bm{L} \bm{L}^\trsp = \left( \int_0^1 \bm{\Phi}_{\textrm{via}}''(s)^\trsp \bm{\Phi}_{\textrm{via}}''(s) ds \right)^{-1},
\end{equation}
that is derived from a probability distribution of smooth trajectories, \ie $p_{\textrm{smooth}}(\bm{q}_{\textrm{via}}, \bm{w}_{\textrm{bc}}) \propto \exp \left( - c_{\textrm{effort}} \right)$, with $c_{\textrm{effort}} = \frac{1}{2} \int_0^1 \bm{q}''(s)^\trsp \bm{q}''(s) ds$.

\subsection{Impedance Control}
\label{sec:control}

At a lower control level, we deploy an impedance controller that runs at a control rate of $1$ kHz, which requires a finely sampled reference trajectory.
Due to our time-continuous representation of the optimized trajectory, we can sample configurations from it with arbitrarily small temporal resolution.
Each MPC step yields an optimized trajectory $\bm{\xi}^*_i$, from which we extract a position-, velocity- and acceleration-reference enabling the robot to track the current movement plan.

\section{Experiments}
\label{sec:exp}

We evaluate the effectiveness and performance of the \emph{VP-STO} framework in simulation, as well as in real-world experiments with a Franka Emika robot arm. 
\subsection{Simulation}
\begin{figure}[t!]
\centering
\includegraphics[width=\linewidth]{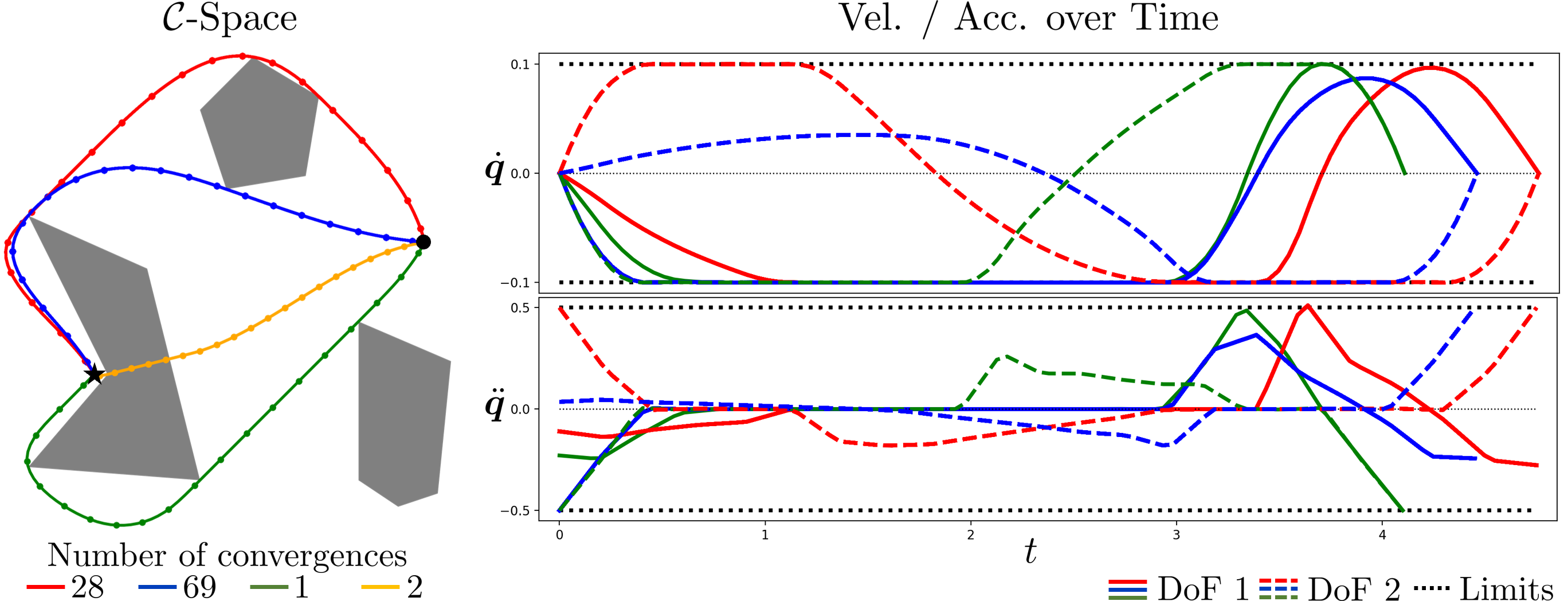}
\caption{\textbf{\emph{Offline VP-STO.}} \emph{Left:} The resulting trajectories from 100 experiment runs when initializing with a straight-line guess between the start position (black circle) and the target position (black asterisk). The number of convergence indicates how often \textit{VP-STO} converged to the corresponding color-coded solution. \emph{Right:} The velocity and acceleration profiles for each degree of freedom corresponding to the valid solutions on the left.} 
\label{fig:planning}
\end{figure}

We begin by evaluating our framework in an offline planning setting for a 2D point mass in a cluttered toy environment adopted from \cite{Bhardwaj2021}. In this experiment, we run \emph{VP-STO} (cf. Sec. \ref{sec:vpsto_offline}) for 100 times with a straight-line initialization. The left plot in Fig. \ref{fig:planning} shows the resulting 100 trajectories after convergence. 
The majority of the found solutions converged to 3 valid local optima, \ie 28 solutions to the red, 69 to the blue and one to the green trajectory. Only 2 runs produce a non-valid solution, shown in yellow. We note here that gradient-based trajectory optimization methods given the straight-line initial guess in such a challenging environment would only converge to this non-valid local optimum. 
Moreover, this also shows that the choice of \mbox{CMA-ES} as a solver for our framework helps to converge to the present local optima with negligible error, despite the stochasticity in the sampling of the via-points. Last, the corresponding velocity and acceleration profiles (only shown for the valid solutions), depicted on the right of Fig. \ref{fig:planning}, reflect the timing-optimal property of the generated trajectories.
After applying maximum acceleration at the start of the movement, the robot moves at maximum speed within the limits before it again applies the maximum acceleration to stop at the goal.
This implies that our framework generates trajectories that not only respect the given dynamic limits, but also exploits them in the spirit of timing-optimality. 

\begin{figure}[t!]
\centering
\includegraphics[width=\linewidth]{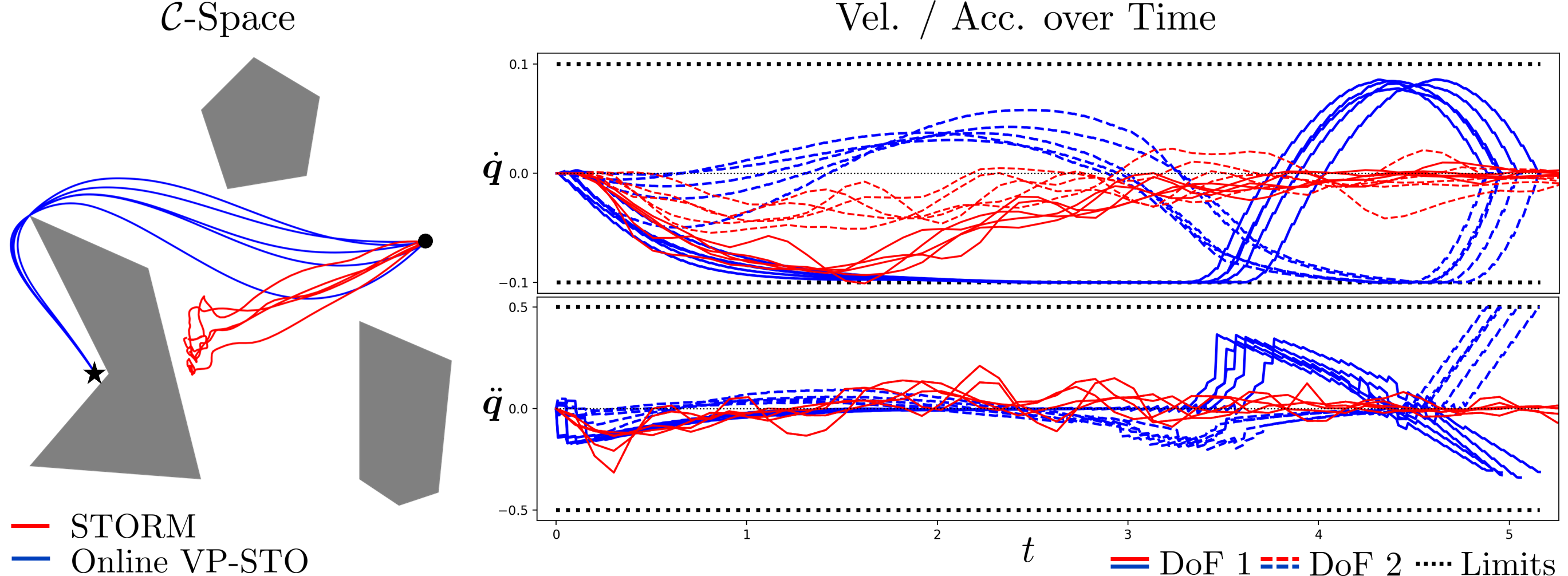}
\caption{\textbf{\emph{Online VP-STO (MPC).}} \emph{Left:} The trajectories taken by the robot when deploying \emph{VP-STO} in an MPC setting (blue), as opposed to using \emph{STORM} \cite{Bhardwaj2021} (red). \emph{Right:} The velocity and acceleration profiles for each degree of freedom corresponding to the found solutions on the left.}
\label{fig:mpc}
\vskip -16pt
\end{figure}

For the online setting, as described in Sec. \ref{sec:vpsto_online}, we compare \emph{VP-STO} to \emph{STORM} \cite{Bhardwaj2021}, which we consider as state-of-the-art in sampling-based MPC for producing reactive robot behavior. Again using the scenario from above, we run 5 experiments in which we deploy \emph{VP-STO} within the MPC-algorithm (cf. Alg. \ref{alg:mpc}). The resulting trajectories are shown in blue in Fig. \ref{fig:mpc} alongside the 5 solutions in red generated by \emph{STORM}. It can be seen that \emph{STORM} is not able to reach the goal. Especially, due to the short-horizon optimization scheme, the robot first follows the path with the shortest distance towards the goal while not being able to anticipate moving around the obstacle early enough. Therefore, it gets stuck in front of the obstacle. In contrast, \emph{Online VP-STO} produces solutions which allow the robot to smoothly navigate to the goal, while exploiting its velocity and acceleration limits. The given setting and experiment emphasizes the advantage of our efficient formulation which allows us to always optimize over the full horizon. 

\subsection{Real-World Experiments}
We demonstrate \emph{VP-STO} on a real robot using the manipulation scenarios in Fig.~\ref{fig:panda_exp}: a \emph{pick-and-place} and a \emph{box pushing} task. We increase the complexity of both scenarios by disturbing the robot and the target objects. This requires a fast feedback loop provided by \emph{Online VP-STO}.

\vspace*{1ex}
\noindent\textbf{Setup.}
Both experiments are performed on a Franka Emika robot arm.
The framework was run on Ubuntu 20.04 with an Intel Core i7-8700 CPU@3.2GHz and 16GB of RAM.  The poses of the objects were tracked with a Vicon motion capture system and post-processed with an extended Kalman filter.
The MPC steps are executed at a fixed control rate (specified below).
In a single MPC step, we run optimization iterations until the next MPC step starts.
%


\paragraph{Pick-and-Place}
First, we consider a \emph{pick-and-place} scenario under human intervention. The robot's task is to grasp a pin, \ie the picking phase, and to place it in an upright position in a given target location in the workspace, \ie the placing phase.
In the picking phase, the pin can be either handed over to the robot in arbitrary poses or the robot needs to pick it up from the table. This phase requires real-time collision avoidance in narrow configuration passages, \ie the robot has to avoid collisions between its hand, including the fingers, and the pin while reaching a configuration where the hand encloses the pin. For the grasp pose, we run a separate pose optimization process in parallel to \textit{VP-STO}, providing the final robot configuration $\bm{q}_T$.
After a successful grasp, the robot continues with the placing phase. 
The challenge here is that the pin might still move within the gripper due to its own weight or due to interference from a user. 
Consequently, feedback of the current pin pose is needed to avoid collisions between the pin and the environment and to correctly place the pin.
We parameterize the sampled trajectories with a maximum number of via-points $N_{\textrm{max}}{=}4$ and $\alpha{=}2$.
\textit{VP-STO} replans with a rate of $12.5$ Hz.

\paragraph{Box Pushing}
In the second scenario, we address the task of planning and control through physical contacts, \ie the robot is supposed to push a box towards a moving target position.
Such a task requires the robot to deliberately make and break contacts, which is subject to discontinuous cost-landscapes.
Here, we exploit the presented trajectory parameterization by setting the final robot configuration $\bm{q}_T$ of each MPC step such that the end-effector moves towards the center of the box.
This enforces all sampled candidate trajectories to make contact with the box. The point of contact and the resulting dynamics of the box depends on the location of the via-points which are subject to minimizing the distance between the box position and the target.
For the sake of fast simulations of the contact dynamics, we use a quasi-dynamic model for the box dynamics parallel to the table surface.
\textit{VP-STO} is executed with a constant number of via-points $N{=}3$ at a control rate of $20$~Hz.


\vspace*{1ex}
\noindent\textbf{Cost Terms.}
We begin with the task-agnostic terms and conclude with more task-specific terms. 
\paragraph*{Movement Duration}
The movement duration is used explicitly as part of the cost function in order to minimize the time needed for the remaining robot movement.
\paragraph*{Smoothness}
In order to optimize not only for fast, but also efficient movements, we use the same metric as in Eq.~\eqref{eq:obf} as the smoothness cost term.
\paragraph*{Joint Limit Avoidance}
For keeping the robot configuration inside the joint angle limits, we deploy a discontinuous metric that accounts for joint limit violations, \ie
\begin{equation}
c_{\textrm{jla}}(q) = 
\begin{cases}
    1 + q - q_{\max},& \text{if } q \geq q_{\max}\\
    1 + q_{\min} - q,& \text{if } q \leq q_{\min}\\
    0,              & \text{otherwise}
\end{cases}.
\end{equation}
We consider a trajectory to be invalid if it results in a joint limit violation, \ie $q \geq q_{\max}$ or $q \leq q_{\min}$.
\paragraph*{Collision Avoidance}
In order to efficiently evaluate the validity of a trajectory regarding collisions between the robot and the environment, we perform binary collision checks for each configuration evaluated along the trajectory, instead of computing a distance between two geometries. 
Thus, the collision cost for a single trajectory is equal to the number of evaluation points that are in collision.
Similarly to the joint limit avoidance cost, we consider a trajectory to be invalid if it results in a collision.
\paragraph*{Pushing Progress}
In the case of a pushing task, we further require a cost term that rewards trajectories which let the robot move the box closer to the current desired target $\bm{x}_{\textrm{box}}^{\textrm{des}}$.
We evaluate the pushing progress of a single trajectory by first simulating the contact dynamics that result in a trajectory of the box $\bm{x}_{\textrm{box}}(t)$; and then computing the box position error at the beginning $e_{\textrm{box}, 0} = || \bm{x}_{\textrm{box}}(0) - \bm{x}_{\textrm{box}}^{\textrm{des}} ||_2^2$ and at the end $e_{\textrm{box}, T} = || \bm{x}_{\textrm{box}}(T) - \bm{x}_{\textrm{box}}^{\textrm{des}} ||_2^2$ of the robot movement.
The final pushing progress cost is given by $c_{\textrm{push}}(\bm{\xi}) = \exp(e_{\textrm{box}, T} - e_{\textrm{box}, 0})$.
Additionally, we consider trajectories that move the box away from the target, \ie $e_{\textrm{box}, T} \geq e_{\textrm{box}, 0}$, to be invalid. In that case, the \textit{exploration} mode in the next MPC step (cf. Sec.~\ref{sec:init}) is triggered.

\vspace*{1ex}
\noindent\textbf{Results.}
First, we note that throughout the experiments, the robot did not collide with any objects in the workspace and did not violate the joint limits.
When the experimenter perturbs the robot, \ie disturbing it through physical interaction or pulling the pin out of the gripper, the robot is compliant and adapts its motion. 
In the pick-and-place scenario, it robustly picked up the pin from various locations in the workspace, including handovers by the experimenter; and placed it at the desired target location in all runs.
In the box pushing scenario, the robot manages to find pushing motions from arbitrary configurations and box locations and to eventually push the box into the target.
We note, however, that some changes of the target location resulted in the robot not finding a valid pushing motion quickly enough, which in turn made the robot push the box out of the workspace. This could only be recovered by the experimenter.
Recordings of the experiments and additional material can be found in the accompanying video to this paper and on the dedicated website \url{https://sites.google.com/oxfordrobotics.institute/vp-sto}.


\section{Conclusion}

We presented a motion optimization framework that is able to generate reactive and yet smooth and efficient robot behavior for complex high-dimensional robot tasks. In contrast to standard trajectory optimization techniques, sampling-based and gradient-based, our framework outputs trajectories which not only optimize over space but also time.
Moreover, due to the full-horizon optimization in an MPC-setting, it is particularly suitable for closed-loop manipulation tasks that demand for continuous re-planning and feedback. We successfully demonstrate this in two real-world experiments on a Franka Emika robot arm, \ie a pick-and-place and a box-pushing scenario.

We wish to extend and improve our work by considering the following points. First, the number via-points to sample yet is subject to heuristic tuning. In general, with increasing movement complexity more via-points are needed at the cost of higher computational complexity. Future work should make the selection of this hyper-parameter more intuitive. 
And second, we would like to further increase the robustness of \textit{VP-STO} by considering uncertainties in the interaction between the robot and its environment.
This includes to explore stochastic roll-outs in the cost evaluation. 

\newpage

%
\bibliographystyle{IEEEtran}
\bibliography{refs_clean}


\clearpage
\section*{APPENDIX}

\subsection{Efficient Gaussian Sampling of Smooth Trajectories through Covariance Matrix Decomposition (Extended)}

In the proposed MPC algorithm, the number of optimization iterations ran in a single step is limited by the desired control rate and the computational resources.
We have identified two modifications of the algorithm that drastically reduce the cost of the mean trajectory after a given optimization time budget.

First, we replace the standard CMA-ES optimization by sep-CMA-ES, a variant that iterates only on diagonal covariance matrices.
This improves the computational complexity from $\mathbb{O}(N^2 D^2)$ (CMA-ES) to $\mathbb{O}(N D)$ (sep-CMA-ES), meaning that the computational load of sampling from and updating the covariance matrix scale linearly with the number of via-points $N$ and the DoF $D$ of the robot.

Second, instead of initializing the covariance matrix $\bm{\Sigma}_{\mathrm{via}}$ with an identity matrix scaled by a single scalar, we start the optimization with a covariance matrix that captures smoothness correlations between via-points.
This modification can be justified by a probabilistic view on stochastic optimization problems, \ie rather than minimizing the expected cost $c(\bm{q}_{\textrm{via}})$ as in \eqref{eq:cost_func}, we aim at maximizing a probability $p(\bm{q}_{\textrm{via}}) \propto e^{ - c(\bm{q}_{\textrm{via}})}$.
It is easy to show that both optimization problems have equivalent optima.
In fact, CMA-ES attempts to locally approximate the generally intractable probability distribution $p(\bm{q}_{\textrm{via}})$ by a Gaussian distribution in each iteration.
If the trajectory cost is given as a sum of multiple objectives, \ie $c(\bm{q}_{\textrm{via}}) = \sum_i c_i(\bm{q}_{\textrm{via}})$, the corresponding probability distribution can be written as a product of multiple probability distributions, \ie $p(\bm{q}_{\textrm{via}}) \propto \prod_i e^{ - c_i(\bm{q}_{\textrm{via}})}$.
A smoothness metric is typically part of the cost function, in our case we use
\begin{align}
\begin{split}
\label{eq:smoothness}
  c_{\textrm{smooth}}(\bm{q}_{\textrm{via}}) &= \frac{1}{2} \int_0^1 \bm{q}''(s)^\trsp \bm{q}''(s) ds\\
  &= \frac{1}{2} \bm{w}^\trsp  \int_0^1 \bm{\Phi}''(s)^\trsp \bm{\Phi}''(s) ds \bm{w}.
\end{split}
\end{align}
Since $\bm{w} = \left[ \bm{q}_{\textrm{via}}^\trsp, \bm{w}_{\textrm{bc}}^\trsp \right]^\trsp$, the smoothness cost term can be exactly represented by a Gaussian distribution, \ie
\begin{equation}
    p_{\textrm{smooth}}(\bm{q}_{\textrm{via}}, \bm{w}_{\textrm{bc}}) = \mathcal{N}\left(\bm{0}, \int_0^1 \bm{\Phi}''(s)^\trsp \bm{\Phi}''(s) ds\right).
\end{equation}
We condition the joint distribution on the given boundary constraints to obtain the corresponding distribution of the via-points $p_{\textrm{smooth}}(\bm{q}_{\textrm{via}} | \bm{w}_{\textrm{bc}}) = \mathcal{N}(\bm{\mu}_{\textrm{via, smooth}}, \bm{\Sigma}_{\textrm{via, smooth}})$ with
\begin{align}
\begin{split}
\label{eq:smoothness2}
  &\bm{\mu}_{\textrm{via, smooth}} = \bm{\Sigma}_{\textrm{via, smooth}} \int_0^1 \bm{\Phi}_{\textrm{via}}''(s)^\trsp \bm{\Phi}_{\textrm{bc}}''(s) ds \; \bm{w}_{\textrm{bc}} \\
  &\bm{\Sigma}_{\textrm{via, smooth}} = \left( \int_0^1 \bm{\Phi}_{\textrm{via}}''(s)^\trsp \bm{\Phi}_{\textrm{via}}''(s) ds\right)^{-1}.
\end{split}
\end{align}
By initializing the covariance matrix with $\bm{\Sigma}_{\textrm{via, smooth}}$, the very first population of via-points that is evaluated in an optimization loop is consequently sampled from $p_{\textrm{smooth}}$.
This can be interpreted as an informed warm-starting of the covariance matrix in a CMA-ES loop.

In order to integrate the off-diagonal structure of $\bm{\Sigma}_{\textrm{via, smooth}}$ with the diagonal covariance matrix $\mathrm{diag}(\bm{\sigma}_{\mathrm{via}})$ that is updated by sep-CMA-ES, we assemble the final covariance matrix by a Cholesky factorization, \ie $\bm{\Sigma}_{\mathrm{via}} = \bm{L} \mathrm{diag}(\bm{\sigma}_{\mathrm{via}}) \bm{L}^\trsp$.
The off-diagonal structure is imposed by the lower triangular matrix $\bm{L}$ that is given by the Cholesky decomposition of the smoothness covariance, such that $\bm{\Sigma}_{\textrm{via, smooth}} = \bm{L} \bm{L}^\trsp$.






\begin{figure*}[ht!]
    \centering
    \includegraphics[width=.8\linewidth]{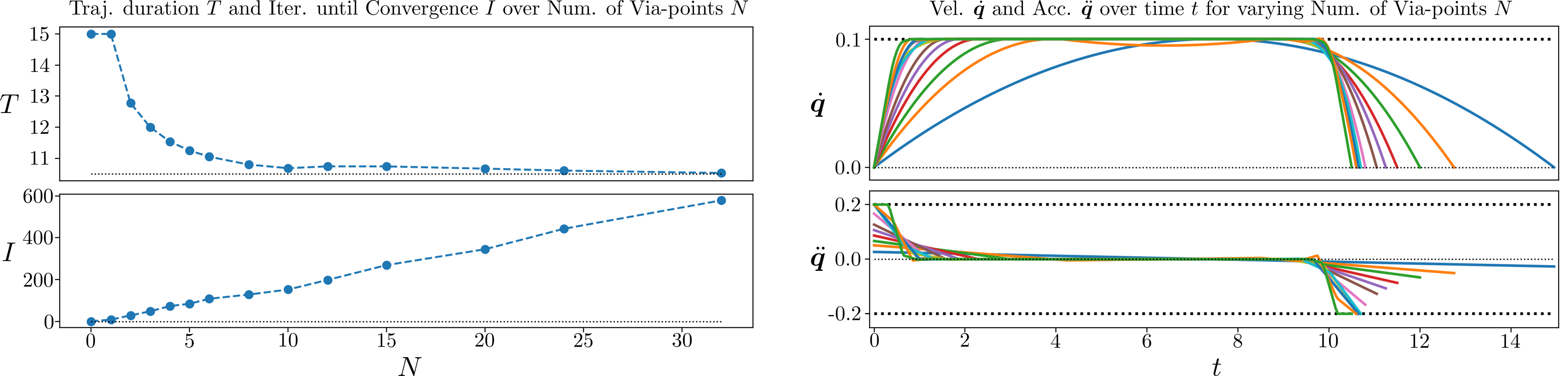}
    \caption{A study of the impact of the number of via-points in a 1D time-optimization problem. \textbf{Top-Left:} Impact on the resulting movement duration. The dotted black line illustrates the duration of the optimal \textit{bang-bang} solution. \textbf{Bottom-Left:} Impact on the number of iterations required until convergence. \textbf{Right:} Velocity and acceleration profiles for evaluated numbers of via-points.}
    \label{fig:ab_via}
\end{figure*}

\subsection{Ablation Studies}

In the paper, we present design choices that we want to further justify via ablation studies.

\subsubsection{Impact of the Number of Via-points}

In this ablation study, we investigate the impact of the number of via-points used to represent the robot movement.
This hyper-parameter has a high impact on the overall framework performance.
On one hand, it directly sets the dimensionality of the optimization problem to solve; on the other hand, it directly spans the space of movements that can be synthesized.
From an optimization perspective, tuning the number of via-points gives us an intuitive way of increasing/decreasing resources on an optimization result with a decreasing/increasing cost.
We illustrate this relationship in Fig.~\ref{fig:ab_via}, where we let a 1D double-integrator move from $q_0 = 0.0$, $\dot{q}_0 = 0.0$ to $q_T = 1.0$, $\dot{q}_T = 0.0$ in minimal time, subject to a maximum velocity $|\dot{q}| < 0.1$ and an acceleration limit $|\ddot{q}| < 0.2$; with a varying number of via-points.
This time-optimal control problem is known to be solved by a bang-bang acceleration profile, such that we know the analytic limit of the minimal time to be $c_{\textrm{bang-bang}} = T_{\textrm{bang-bang}} = 10.5$, which is depicted as dashed black line in the upper-left plot.
We observe that the solution cost exponentially converges to $c_{\textrm{bang-bang}}$ as we increase the number of via-points.
The lower-left plot shows the number of CMA-ES-iterations required to converge as a function of the number of via-points.
Here, we detect convergence if $|c_k - c_{k-1}| < 10^{-6}$ in the $k$-th iteration.
Interestingly, the number of iterations grows linearly with the number of via-points.
Note that this does not mean that the computational cost grows linearly with the number of via-points, since the computational cost for a single iteration is either linear (sep-CMA-ES) or quadratic (CMA-ES) in the number of via-points.
Nevertheless, those results motivate to use a low number of via-points as with a growing number of via-points, the benefit of adding a via-point is not worth the extra computational cost. 

\subsubsection{Impact of the Cholesky Factorization of the Covariance Matrix}

\begin{figure*}[t!]
    \centering
    \includegraphics[width=\linewidth]{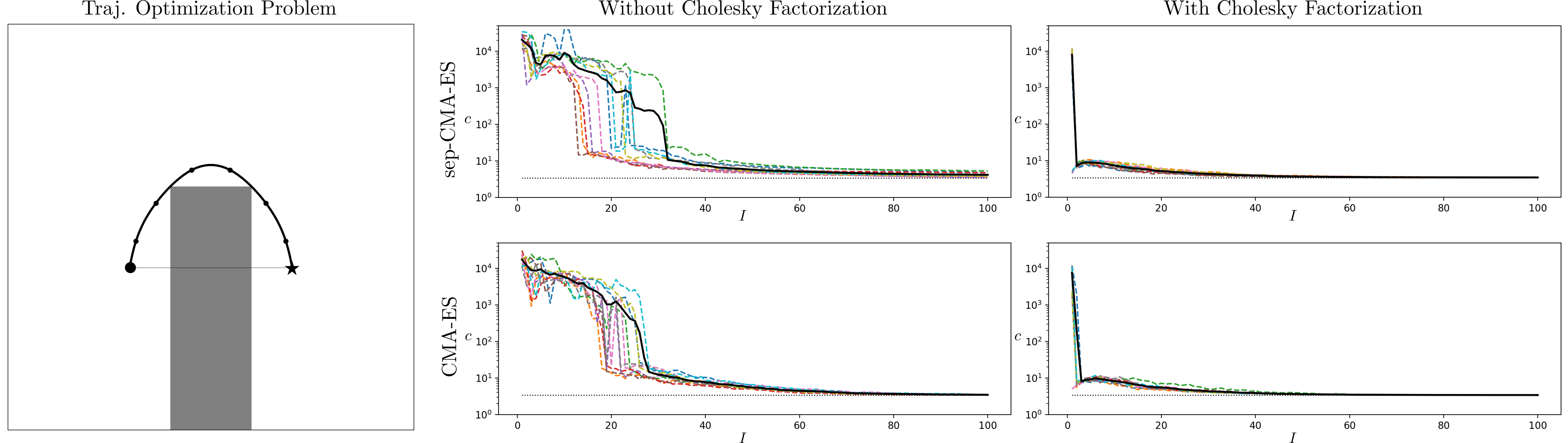}
    \caption{A study of the impact of the Cholesky factorization of the Covariance Matrix $\bm{\Sigma}_{\mathrm{via}}$ in a 2D time-optimization problem with obstacle avoidance. \textbf{Left:} The configuration space including the obstacle in gray, the initial guess as dashed line, and the optimal solution around the obstacle as solid line together with the corresponding via-points as circles. \textbf{Center:} The via-point covariance matrix is explicitly updated, \ie $\bm{\Sigma}_{\mathrm{via}} = \bm{\Sigma}_{\mathrm{CMA}}$. \textbf{Right:} The via-point covariance matrix is updated through a Cholesky factorization, \ie $\bm{\Sigma}_{\mathrm{via}} = \bm{L} \bm{\Sigma}_{\mathrm{CMA}} \bm{L}^{\trsp}$. \textbf{Top:} sep-CMA-ES iterates on diagonal covariance matrices only, \ie $\bm{\Sigma}_{\mathrm{CMA}} = \textrm{diag}(\bm{\sigma}_{\mathrm{CMA}})$, with linear computational complexity $\mathcal{O}\left(ND\right)$. \textbf{Bottom:} CMA-ES iterates on full covariance matrices $\bm{\Sigma}_{\mathrm{CMA}}$ with quadratic computational complexity $\mathcal{O}\left(N^2D^2\right)$.}
    \label{fig:ab_Chol}
\end{figure*}

In this ablation study, we look at a 2D minimal-time planning problem including an obstacle that is to be avoided.
We fix the number of via-points to $N=6$ and set up four different optimization loops that are supposed to solve the same problem.
Each setup uses either CMA-ES or sep-CMA-ES and runs with or without the Cholesky factorization of the covariance matrix as described in Sec.~\ref{sec:sampling}.
For comparison, we look at the cost evolution over the number of iterations.
The dashed black line in all plots (except for the left-hand plot) indicates the minimum cost measured in any experiment.
Note also the jump in all the cost profiles from $\approx 10^3 - 10^4$ to $\approx 10^0 - 10^1$, which reflects if the updated solution is collision-free.
We observe that the choice of CMA-ES vs. sep-CMA-ES does not have a substantial impact on the cost evolution for this particular problem, indicating that it is justified to use sep-CMA-ES with linear complexity.
However, we observe a substantial impact when using the presented Cholesky factorization, imposing smoothness on the candidate trajectories.
In all experiments using the Cholesky factorization, it converged to a collision-free solution after 3 iterations at maximum.
This is an especially important result justifying the use of the Cholesky factorization inside the MPC loop, as the real-time requirements limit the number of iterations.




\end{document}